\documentclass[default,iicol]{sn-jnl}

\jyear{2021}%

\theoremstyle{thmstyleone}%
\newtheorem{theorem}{Theorem}
\newtheorem{proposition}[theorem]{Proposition}%

\theoremstyle{thmstyletwo}%
\newtheorem{example}{Example}%
\newtheorem{remark}{Remark}%

\theoremstyle{thmstylethree}%
\newtheorem{definition}{Definition}%

\usepackage{romannum}
\usepackage{graphicx}
\usepackage{amsmath}
\usepackage{amsthm}
\usepackage{multirow}
\usepackage{xcolor}

\renewcommand{\set}[1]{\{#1\}}
\newcommand{\automaton}[1]{\langle #1\rangle}
\newcommand{\word}[1]{\automaton{#1}}
\newcommand{\tabincell}[2]{\begin{tabular}{@{}#1@{}}#2\end{tabular}}

\begin{document}

\title[Anomaly Detection in DNN]{A Uniform Framework for Anomaly Detection in Deep Neural Networks}


\author[1]{\fnm{Fangzhen} \sur{Zhao}}

\author[1]{\fnm{Chenyi} \sur{Zhang}}

\author[2]{\fnm{Naipeng} \sur{Dong}}

\author[1]{\fnm{Zefeng} \sur{You}}

\author[1]{\fnm{Zhenxin} \sur{Wu}}




\affil[1]{\orgdiv{College of Computer Science and Technology}, \orgname{Jinan University}, \orgaddress{\city{Guangzhou}, \country{China}}}

\affil[2]{\orgdiv{School of Information Technology and Electric Engineering}, \orgname{University of Queensland}, \orgaddress{\city{Brisbane}, \state{Queensland}, \country{Australia}}}


\abstract{Deep neural networks (DNN) can achieve high performance when applied to In-Distribution (ID) data which come from the same distribution as the training set. When presented with anomaly inputs not from the ID, the outputs of a DNN 
should be regarded as meaningless. However, modern DNN often predict anomaly inputs as an ID class with high confidence, which is dangerous and misleading.
%
In this work, we consider three classes of anomaly inputs, (1) natural inputs from a different distribution than the DNN is trained for, known as Out-of-Distribution (OOD) samples, (2) crafted inputs generated from ID by attackers, often known as adversarial (AD) samples, and (3) noise (NS) samples 
generated from meaningless data. We propose a framework that aims to detect all these anomalies for a pre-trained DNN. Unlike some of the existing works, our method does not require preprocessing of input data, nor is it dependent to any known OOD set or adversarial attack algorithm. Through extensive experiments over a variety of DNN models for the detection of aforementioned anomalies, we show that in most cases our method outperforms state-of-the-art anomaly detection methods in identifying all three classes of anomalies.}

\keywords{Deep neural networks,  Out-of-Distribution, Adversarial Attack, Anomaly Detection}



\maketitle



\section{Introduction}
Although being an emerging technology, Deep Neural Networks (DNN) is valued to be \$38.71 billion globally by 2023, with wide range of applications cross sectors like finance, energy \& utilities, retail, IT \& telecom, manufacturing, aerospace \& defence, healthcare etc. (according to Allied Market Research)~\cite{AMR}.
Along with DNN's popularity is a growing concern on the safety of DNN models in carrying out the tasks (typically classification) in those applications, especially the security or safety critical ones such as healthcare and self-driving vehicles.

To address the concern, the foremost issue is to ensure the quality of the input data, which the DNN models depend on. 
As a data-driven technique, a DNN model will only be as good or as bad as the data provided (training data). 
For instance, a speech recognition system trained on clean speech will not perform well on noisy speech. 
However, when applied to real-world tasks, it is inevitable that the testing data differs from the training data due to a variety of reasons, such as mis-operations in data collection, natural noise, untrusted data resources etc. Such unavoidable anomaly testing data leads to severe safety problems - the DNN tends to provide high-confidence predictions while being woefully incorrect~\cite{baseline}.

A variety of works exist aiming to detect anomaly in testing data, most of which focus on the applications of image classification. We classify these works according to the types of anomaly data they handle: out-of-distribution (OOD) data, adversarial (AD) data, and noise (NS) data.

\begin{description}
\item[\textbf{OOD}]
Out-of-Distribution (OOD) data refers to inputs that do not contain any of the classes modeled in the 
training distribution. For example, the clothing images from Fashion-MNIST are OOD data for a DNN trained with the MNIST data set which consists of hand-written digits. The OOD data considered in this paper are collections of \emph{meaningful} \emph{natural} images that are not from ID, excluding those \emph{crafted} images (known as adversarial data) and \emph{meaningless} images (classified as noise data).
\item[\textbf{AD}]
Adversarial (AD) data is generated by introducing an imperceptible perturbation to an image from the in-distribution (ID), with the intention of inducing a DNN to make wrong judgments. The adversarial methods used in our experiments include FGSM~\cite{FGSM}, BIM~\cite{KurakinGB16}, JSMA~\cite{PapernotMJFCS15} and CW~\cite{cw}. 
\item[\textbf{NS}]
We consider two types of noise (NS) data. The first type (NS-\Romannum{1}) is merely random noise (e.g., Gaussian noise). The second type (NS-\Romannum{2}) is often known as fooling images, which are created by evolving meaningless images in order to mislead a DNN to output classes in ID with high confidence, such as the images generated in~\cite{nguyen2015deep} (examples shown in Figure~\ref{fig:noise2}).
\end{description}

Through a comprehensive literature study (see Table~\ref{table:related-work} in Section~\ref{sec:related_work}), we observe that existing works either focus only on OOD and NS-\Romannum{1} detection, or focus on AD detection. Few works detect both OOD and AD (a notable exception is~\cite{Mahalanobis}); and no works aim to detect NS-\Romannum{2}.

As suggested by~\cite{ADNotDetec2017}, there 
are no known intrinsic properties that differentiate %
natural images and adversarial images. 
We believe that %
for real-world tasks, 
all types of anomalies could potentially be fed into DNN models,
and there is no effective way to tell ahead of time if they are in-distribution, adversarial, or out-of-distribution images%
. Many existing works make an implicit assumption that the analysers know which type of anomaly in advance and build detection approaches for the particular type, which is impossible in reality. Therefore, detection approaches should be developed to be able to 
handle all three types of anomalies that people are aware of.

The uncertainty of the anomaly types leads to a non-trivial challenge in the anomaly detection task.
During experiments, we observe that, (\romannum{1}) in general, an approach for OOD performs badly for detecting AD and vice versa. Any combination of the results for each approach would not be able to outperform 
either the best result for OOD or the best result for AD. 
In addition, we observe (\romannum{2}) many existing works require pre-processing to the input data. This hinders their applicability in real-world tasks, since one needs to know, in advance,  which data is ID and which is OOD in the case of OOD detection, and needs to know 
by which adversarial algorithm the data is generated in the case of AD detection. Removing the prior knowledge would largely 
degrade performance. Therefore, this work improves the anomaly detection accuracy without requiring prior knowledge of input data. 
Our approach detects anomaly using features from the Most Discriminative Layer (MDL) which has better distinction for the sub-domain of the test input. The anomaly detector then combines the MDL features with the Logit layer to cover all the three types of anomalies.




\paragraph{Contributions.} This paper proposes a uniform framework for detecting anomaly inputs, which has wider coverage of the anomaly types, easier applicability to various models, and better performance in accuracy, compared to the state-of-the-art approaches.
\begin{description}
\item[Coverage] Our approach is able to detect all three types of anomaly inputs, including OOD, AD and NS, which addresses the challenge of unknown anomaly types in real-world applications.
\item[Applicability] Our approach provides a uniform way for the anomaly detection, and thus can be applied to any existing model without requiring extra pre-processing.
\item[Performance] To evaluate our approach, we conduct extensive experiments and a comprehensive comparison with the state-of-the-art approaches. The experiment results show that our approach outperforms the best results in OOD, AD and NS in most of the cases.
\end{description}

\section{Related work}\label{sec:related_work}

\begin{table}[!t]
    \caption{A summary of related work}\label{table:related-work}
    \centering
    {\footnotesize
    \begin{tabular}{ccccc}
        \hline
        Method   & OOD\& NS-\Romannum{1}    & AD &NS-\Romannum{2} & Pre-proc.  \\
        \hline
          Baseline~\cite{baseline} & $\surd$   & $\times$ & $\times$ & $\times$   \\
          ODIN~\cite{ODIN}       & $\surd$   & $\times$   & $\times$  & $\surd$    \\
          ELO~\cite{OODL}      &$\surd$   & $\times$  & $\times$ & $\times$   \\
          OE~\cite{OutlierExposure}     & $\surd$  & $\times$ & $\times$  & $\times$   \\
          G-ODIN~\cite{GeneralizedODIN}   & $\surd$  & $\times$ & $\times$  & $\surd$  \\
        \hline
         KD+BU~\cite{kd+bu}          & $\times$  & $\surd$ &$\times$  &$\surd$  \\
         LID~\cite{LID}       &$\times$    &$\surd$ & $\times$   &$\surd$  \\
         IF~\cite{InfluenceFunction}    &$\times$   &$\surd$  & $\times$ &$\surd$\\
         FS~\cite{FeatureSqueezing}   &$\times$   &$\surd$ & $\times$ & $\times$ \\
        \hline
         MD~\cite{Mahalanobis}       &$\surd$   &$\surd$  &$\times$   &$\surd$   \\
         \hline
    \end{tabular}
    }
\end{table}

As mentioned in the previous section, existing works either focus on OOD and NS-\Romannum{1} detection, e.g.,~\cite{baseline,ODIN,OODL,OutlierExposure,GeneralizedODIN}, or focus on AD detection, e.g.,~\cite{kd+bu,LID,InfluenceFunction,FeatureSqueezing}. The only work that can detect both is Mahalanobis distance (\emph{MD})~\cite{Mahalanobis}.

\noindent
\textbf{OOD+NS-\Romannum{1}.} The seminal work for OOD detection is known as the \emph{baseline} approach~\cite{baseline}, which observes that the softmax prediction probability of OOD tends to be lower than the prediction probability for correct examples, and thus a threshold over the predicted softmax probability can be used to detect OOD. 
A year later, the work \emph{ODIN} observes that temperature scaling and input perturbation (pre-processing) can enlarge the gap between ID and OOD, and thus can be used to improve the detection performance~\cite{ODIN}. Meanwhile, the pre-processing of ODIN requires access to OOD samples in advance, which is impossible in reality, to fine-tune the degree of perturbation. 
To address this limitation, the work Early Layer Output (\emph{ELO})~\cite{OODL} proposes a one-class classifier trained on the output of an early layer instead of the softmax layer,
which does not need to access to OOD samples. An alternative approach called Generalized ODIN (\emph{G-ODIN})~\cite{GeneralizedODIN} is later proposed, 
which only tunes the ID data instead of OOD data.
Differing from the above works, the work Outlier Exposure (\emph{OE})~\cite{OutlierExposure} fine-tunes the pre-trained model using an auxiliary data set that is selected from a disjoint set of OOD samples. OE includes an additional loss function to minimize the distance between the output distribution produced by the pre-trained model for the auxiliary data set and the uniform distribution. The softmax values are used as scores for anomaly detection. 
Similar to ELO and OE, our approach does not require preprocessing and the use of the OOD samples in training.
Nevertheless, our approach 
extracts data from a specific sub-domain, which 
achieves better separation of ID and OOD (see Figure ~\ref{fig:mnist-10}).

\noindent
\textbf{AD.} A variety of approaches have been proposed to detect adversarial samples from their normal and noisy counterparts. For instance, Feinman~\emph{et~al.} 
trains a logistic regression detector using a distance-based generative learning method called kernel density and Bayesian uncertainty features (\emph{KD+BU})~\cite{kd+bu}.
Ma~\emph{et~al.} proposes an intrinsic character of the adversarial regions - the local intrinsic dimensionality (\emph{LID}), as the confidence score to separate the adversarial samples~\cite{LID}. The \emph{IF} approach demonstrates the correspondence 
between the training data and the classification of the network, which is quantified using the Influence Function, and 
outperforms LID~\cite{InfluenceFunction}. Differing from 
the above works, Xu~\emph{et~al.} applies Feature Squeezing (\emph{FS}) to distinguish the adversarial samples from ID
and does not need special treatment to the input data~\cite{FeatureSqueezing}. These works can only detect AD and are not suitable for detecting OOD and~NS.

Finally, the work \emph{MD} measures the probability density of a test sample and uses the Mahalanobis distance as the confidence score to distinguish OOD and AD from ID. It is the first work that can detect both OOD and AD~\cite{Mahalanobis}. Our work is able to detect an additional type (NS-\Romannum{2}) of anomaly and outperforms MD 
on most of the test data sets. %


\newcommand\Input{\mathcal{I}}
\newcommand\Output{\mathcal{O}}
\newcommand\bd[1]{\mathbf{#1}}
\renewcommand\C{C^f}

\section{Preliminaries}\label{sec:preliminaries}

In this section, we introduce the notions that are necessary to understand the remaining part of the paper.
Let a deep neural network (DNN) of $m$ layers be represented as a function $f:\Input\rightarrow\Output$, where $\Input$ is the input domain and $\Output$ is the domain for the output vectors of length $d$. Given $x\in\Input$, we have $f(x)=\word{o_1,\dots,o_d}\in\Output$, and the final classification chosen by the DNN is 
$\C(x)=\underset{i \in \{1,\cdots,d\}}{argmax} \left(o_{i}\right)$ for  
the index of the largest element in vector $f(x)$. 
For $\ell=1\dots m$, we write $f^\ell(x)$ for the output vector in the feature space of layer $\ell$. In the literature, the second last layer (i.e., $\ell=m-1$, right before the softmax layer) is often called the \emph{logit} layer. 

It is commonly assumed that the training data of DNN $f$ is drawn from the distribution $\Delta$ known as ID. 
we write $\Delta_{in}(x)$ if $x\in\Input$ is from ID given $f$.
%
An anomaly detector $g_f$ for DNN $f$ is a binary classifier, such that given input $x\in\Input$, $g_f(x)$ answers whether $x$ is an anomaly with respect to $f$. Since it is often difficult to define what an \emph{anomaly distribution} is, we focus on the three types of anomalies (i.e., OOD, AD and NS) in our experiments.



Our anomaly detection algorithm is based on a discriminative model known as Support Vector Domain Description (SVDD) by~\cite{Tax2004supportvector}.\footnote{Another important approach for one class classification known as $\nu$-SVC is introduced by Sch\"{o}lkopf et al.~\cite{schoelkopf2001}, which can be shown as equivalent to SVDD when the Gaussian kernel is used~\cite{tax:thesis}.} Similar to the famous Support Vector Machine~\cite{vapnik1995}, SVDD defines support vectors for a sphere shaped decision boundary enclosing the class of objects represented by the (unlabeled) training data with minimal space, as shown in the following formulation. 
\begin{equation}\label{eq:svdd}
  \min\limits_{R,\bd{a},\xi}\ R^2+ \frac{1}{n\nu}\sum_i\xi_i
\end{equation}
\[\mbox{s.t. }\forall i:\quad ||\bd{x}_i-\bd{a}||^2\leq R^2+\xi_i,\quad \xi_i\geq 0\]
The solution of the above constraints provides a center vector $\bd{a}$, radius $R$ and 
slack variables $\xi_i$ such that the target term in Eq.~(\ref{eq:svdd}) is minimized, provided that the square of distance from each training data $\bd{x}_i$ to the center $\bd{a}$ may exceed $R^2$ by at most $\xi_i$. Here $\nu$ is a constant in $(0,1]$ and $n$ is the size of the training set. Intuitively, a smaller $\nu$ gives more weight to the right hand side of target term in Eq.~(\ref{eq:svdd}), which imposes smaller values for $\xi_i$ and larger $R$.
%
%
The solution of Eq.~(\ref{eq:svdd}) allows us to determine if a test input $\bd{z}$ is from the ID  by checking the following condition.
\begin{equation}\label{eq:if-id}
  ||\bd{z}-\bd{a}||^2=(\bd{z}\cdot\bd{z})-2\sum_i\alpha_i(\bd{z}\cdot\bd{x}_i)+\sum_{i,j}\alpha_i\alpha_j(\bd{z}\cdot\bd{x}_i)\leq R^2
\end{equation}
Here $\alpha_i$ ($\alpha_j$) is the Lagrange multiplier associated with the constraint for the $i$-th ($j$-th) training input when solving Eq.~(\ref{eq:svdd}), which is non-zero only if the $i$-th ($j$-th) training input is used as a support vector. Given all inputs (including the test input) only appearing in the form of inner product, it is thus viable to replace the inner products by kernel functions, of which the Gaussian Radial Basis Function (RBF) provides the best performance in practice~\cite{tax:thesis}.
The RBF kernel is given in the following formulation, where the free parameter $s$ controls the spread, or how tight the density is, of the kernel.
\begin{equation}\label{eq:rbf}
  K(\bd{x}_{i}, \bd{x}_{j}) = \exp(-\|\bd{x}_{i} - \bd{x}_{j}\|^{2}/s^2)
\end{equation}

Early-Layer Output (ELO)~\cite{OODL}, the work most related to ours, trains a one-class SVDD classifier using an early-layer output of ID data in a latent space, based on the observation that there exists an early-layer called the Most Discriminative Layer (MDL), such that in this latent space the ID data and OOD data are well separated.

%
%

\section{Our approach}

\begin{figure*}[hpbt]
  \centering
  \includegraphics[scale=0.45]{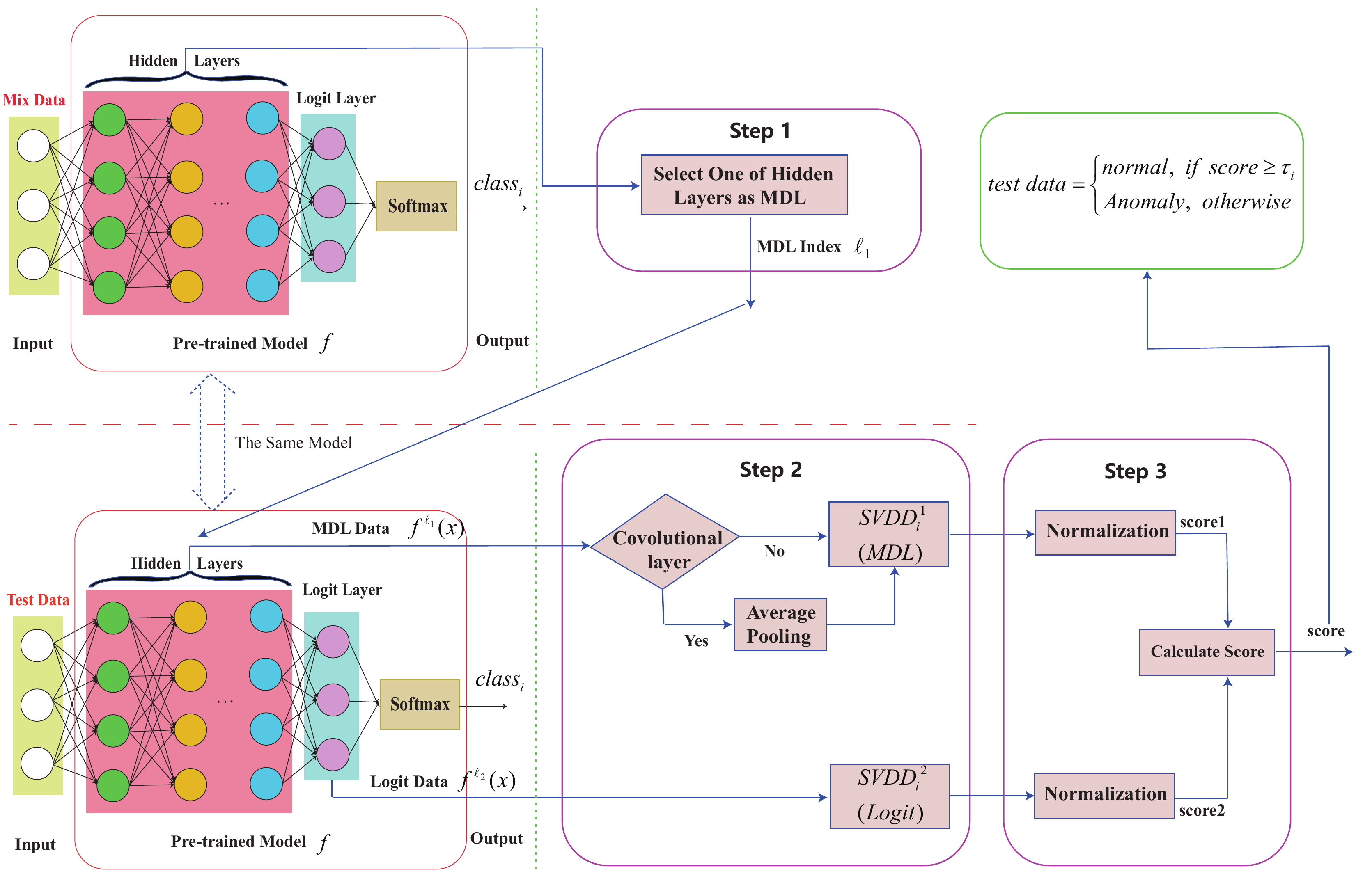}
  \caption{The overview of the proposed approach. The proposed approach has three steps. Step~1: Select one of hidden layers as MDL which has the minimum detection errors for mix data. Step~2: Feed the extracted the data of the MDL and logit layer into the corresponding SVDD$_{i}^{\ell}$. Step~3: Calculate the score of an input sample by using the corresponding SVDD$_{i}^{\ell}$ in MDL and logit layer, which helps to detect anomaly inputs. }\label{fig:overview}
\end{figure*}

We propose a uniform framework for the anomaly detection task. In the training phase, for a given DNN classifier, we empirically choose the Most Discriminative Layer (MDL) using a randomly picked OOD set (mix data), in the way similar to ELO~\cite{OODL}. Then we use the data generated from the MDL layer (Step~1) and the logit layer of the DNN to train two one-class SVDD classifiers for each known class (Step~2). During the testing phase, an input is first given to the DNN classifier which produces an output class $i$. The data from the corresponding MDL and logit layers are forwarded to the corresponding SVDD classifiers, i.e., $\mbox{SVDD}^1_i$ and $\mbox{SVDD}^2_i$, and the scores obtained from the SVDD classifiers are combined to form a final judgment on whether the given input is anomalous (Step~3). The overview of our approach is sketched in Figure ~\ref{fig:overview}.

Let layer $\ell$ be the MDL in a reasonable DNN classifier $f$ to ID with sub-domains for the known classes $1\dots d$, and the set $\set{f^\ell(x)\mid \Delta_{in}(x)}$ forms a manifold of ID in the latent space of layer~$\ell$. We conjecture that anomaly detection precision can be further improved by making predictions conditional to output of the DNN. We focus on the sub-domains $\set{f^\ell(x)\mid \Delta_{in}(x)\wedge \C(x)=i}$ for each known class $i=1\dots d$.

\begin{figure}[hpbt]
  \centering
  \includegraphics[scale=1.0]{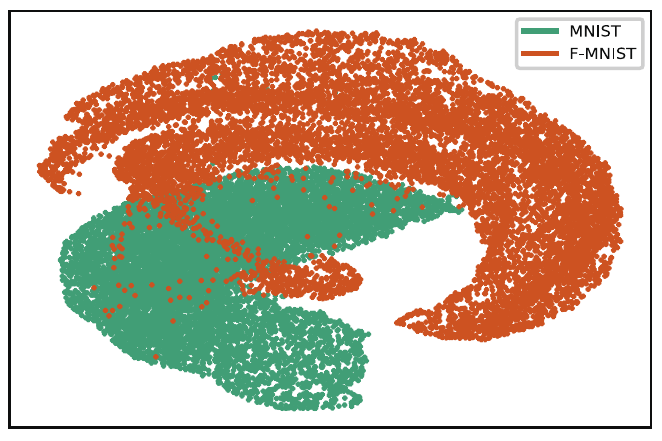}
  \caption{A two-dimensional representation of features extracted from the MDL of a LeNet model trained on MNIST. The feature cluster consisting of data from all $10$ MNIST classes is shown in green dots, and yellow dots represent F-MNIST~(OOD)~data.}\label{fig:mnist}
\end{figure}

\begin{figure*}[th]
  \centering
  \includegraphics[scale=0.9]{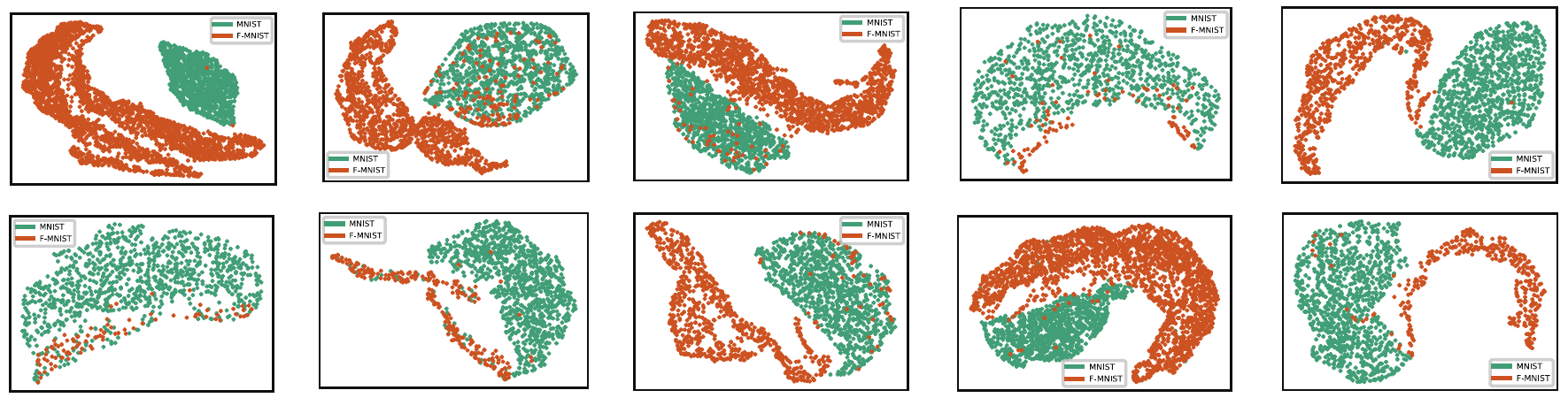}
  \caption{Two-dimensional representations of features extracted from the MDL of a LeNet model trained on MNIST. The feature clusters for the $10$ classes are shown, with green dots for MNIST (ID) data and yellow dots for F-MNIST (OOD)~data.}\label{fig:mnist-10}
\end{figure*}

Taking Eq.~(\ref{eq:svdd}), the approach of ELO empirically chooses $\nu=0.001$, which imposes a strong penalty on samples with distance larger than $R$ (i.e., the slack variables should be very small) when minimizing the entire term of Eq.~(\ref{eq:svdd}), which is reasonable if the sub-domains for different classes are relatively apart. Since $R$ becomes relatively large, potentially more anomaly samples, especially those spatially closer to ID such as adversarial samples, are classified as ID. Moreover, an anomaly input positioned between two ID clusters of distinct classes in the latent space may also be classified as ID.
Figure~\ref{fig:mnist} provides a $2$-D view of the clusters of MNIST samples (green dots) and Fashion-MNIST samples (yellow dots) in the MDL of a LeNet model.\footnote{We choose UMAP~\cite{umap} as the visualization tool.} If we split the MNIST samples and Fashion-MNIST samples into $10$ classes based on the classification results of the LeNet model, and study the MNIST samples and Fashion-MNIST samples confined to each class, we get the $2$-D view in Figure~\ref{fig:mnist-10}, from which it seems that near perfect separation can be achieved in some classes (e.g., the $1$st, $5$th, $9$th and $10$th classes).
Our experiment results shown in Table~\ref{table:OOD} and Table~\ref{table:AD} confirm that the method with sub-domain splitting is comparable (if not marginally better) to the ELO method for the detection of OOD inputs, and for the detection of adversarial inputs the sub-domain based method significantly outperforms ELO.%
\footnote{Empirically, in our experimental setting, we let $\nu=0.1$ which ensures a tighter bound (i.e., a smaller $R$) on each individual class to rule out adversarial samples.}

\begin{figure}[hpbt]
  \centering
  \includegraphics[scale=0.45]{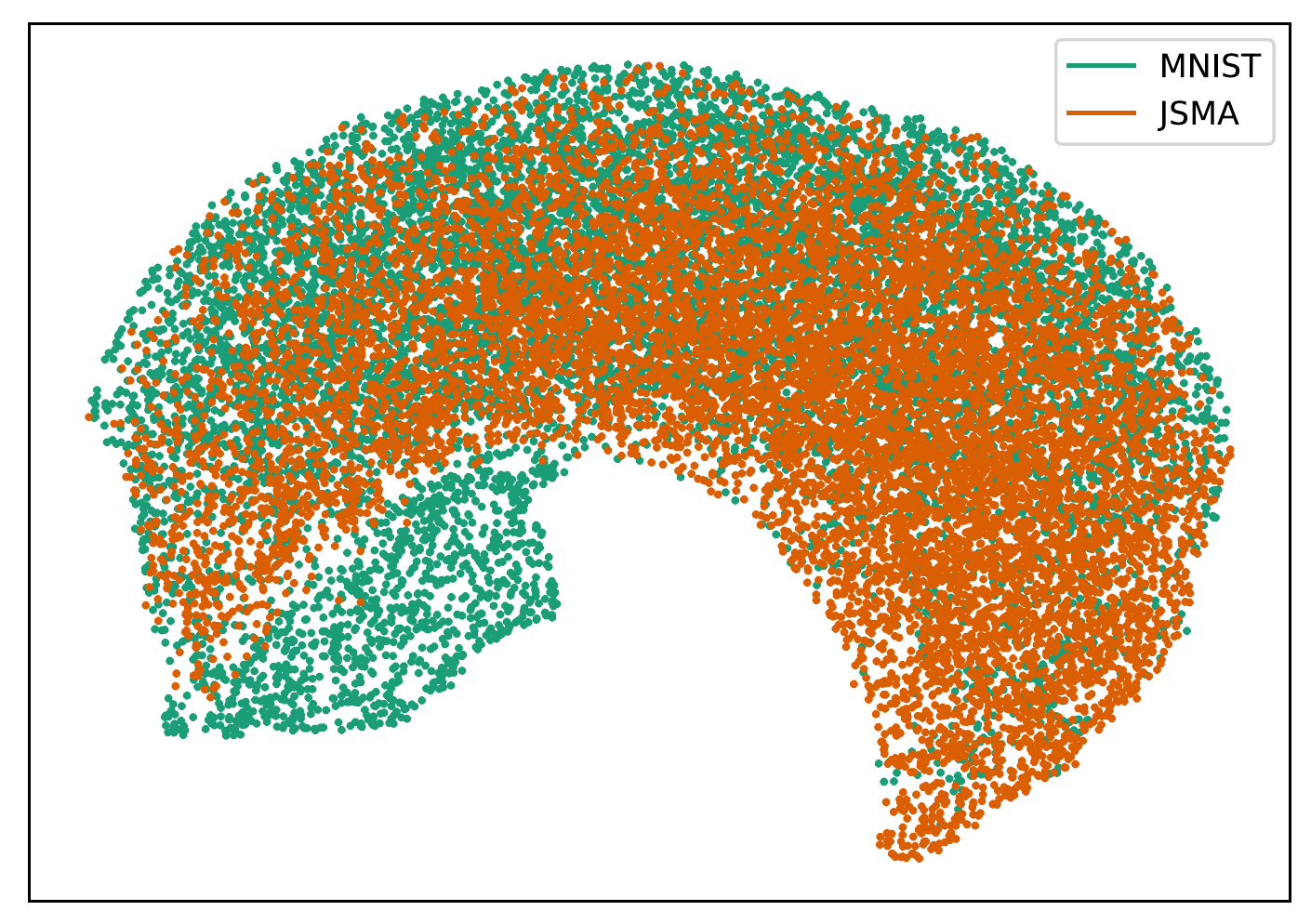}
  \caption{A two-dimensional representation of features extracted from the MDL of a LeNet model trained on MNIST. The feature cluster consisting of all $10$ MNIST classes is shown in green dots, and yellow dots represent JSMA (AD) data.}\label{fig:jsma-1}
\end{figure}

\begin{figure*}[th]
  \centering
  \includegraphics[scale=0.9]{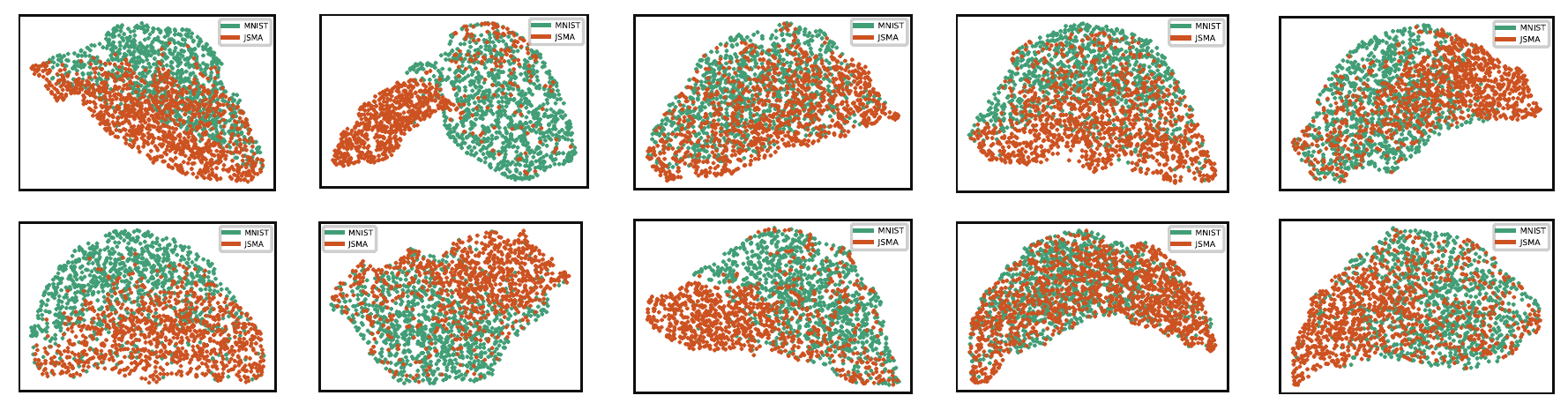}
  \caption{Two-dimensional representations of features extracted from the MDL of a LeNet model trained on MNIST. The feature clusters for the $10$ classes are shown, with green dots for MNIST (ID) data and yellow dots for JSMA (AD) data.}\label{fig:jsma-10}
\end{figure*}

\begin{figure*}[th]
  \centering
  \includegraphics[scale=0.9]{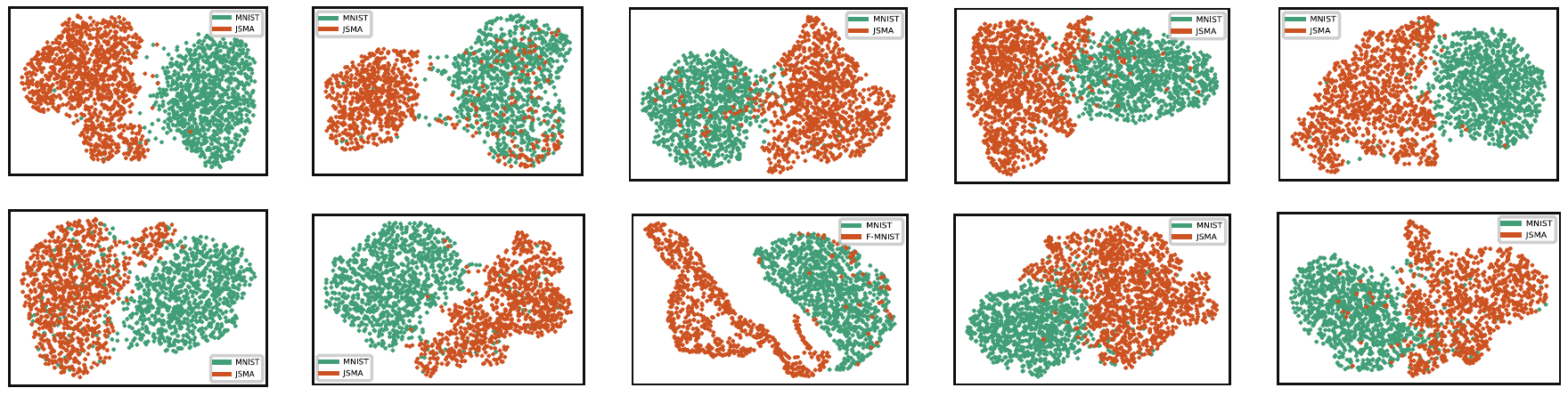}
  \caption{Two-dimensional representations of features extracted from the Logit layer of a LeNet model trained on MNIST. The feature clusters for the $10$ classes are shown, with green dots for MNIST (ID) data and yellow dots for JSMA (AD) data.}\label{fig:jsma-logit-10}
\end{figure*}

Our second observation is that the features represented by different layers of a DNN indeed represent distinctive discriminative power on anomalies. In the literature the softmax and logit layers~\cite{baseline,ODIN,Mahalanobis} are used to distinguish OOD from ID, while some others consider the usage of early layers~\cite{jiang2018trust,OODL}.
Since anomaly data in the class AD are crafted by introducing imperceptible perturbations to images from the ID, they are closer to the ID in the input domain than OOD and NS in the majority of adversarial attack scenarios, especially the AD with relatively small perturbation. Therefore, intuitively, it requires more processing in the original classifier to separate them from ID in the penultimate logit layer, rather than the earlier MDL layer. This 
has also been described in detail in the literature~\cite{DeepKNN}. Figure~\ref{fig:jsma-1} provides a $2$-D view of the clusters of MNIST ID samples (green dots) and JSMA AD samples (yellow dots) in the MDL of a LeNet model. Figure~\ref{fig:jsma-10} and Figure~\ref{fig:jsma-logit-10} show the feature clusters for the $10$ classes of MNIST ID samples (green dots) and JSMA AD samples (yellow dots) in the MDL and logit layer of a LeNet model, respectively. Such a conjecture is also 
confirmed by the results in Table~\ref{table:OOD} and Table~\ref{table:AD}.
We believe that combining the power of the early layers and the late layers can achieve better precision on detecting different types of anomalies.

\begin{figure*}[th]
  \centering
  \includegraphics[scale=1.05]{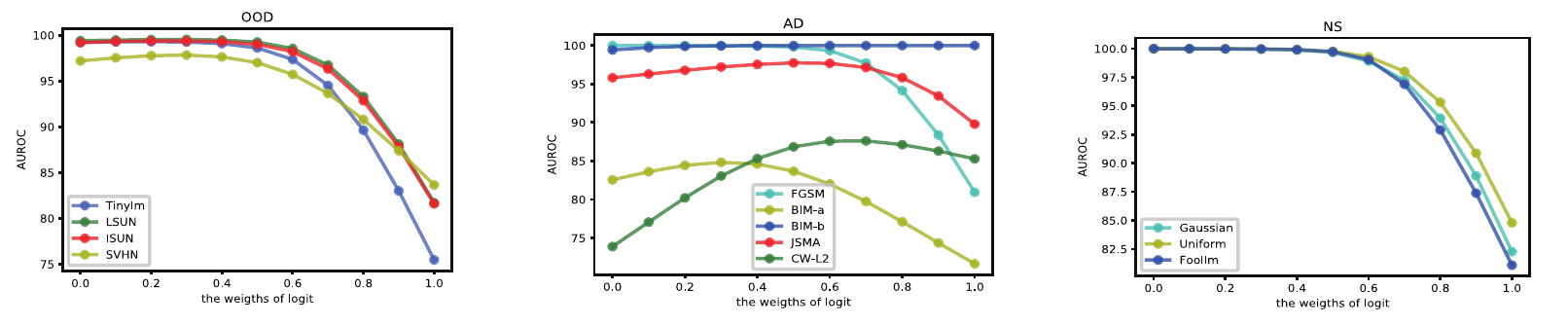}
  \caption{AUROC of the combined anomaly detection model with different weight coefficients for values from the logit layer of a ResNet model (CIFAR-10 as ID)}\label{fig:coefficients}
\end{figure*}

Based on the above observations, we train 
two SVDD detectors for
layers $\ell_1$ and $\ell_2$ for each class $i\in\set{1,\dots d}$, and combine the results by defining $g_i(x) = \beta_1\cdot g^{\ell_1}_{f,i}(x)^* + \beta_2\cdot g^{\ell_2}_{f,i}(x)^*$ as a score to determine if $x$ is an anomaly, given $\C(x) = i$. In the above formulation, we choose the MDL as $\ell_1$ 
which is empirically determined and it is 
most of the time an early layer that gives better precision on detection anomalies than any other layers, 
and $\ell_2$ is the logit layer. When used for combining scores, $g^{\ell}_{f,i}(x)^*$ is the normalized value of $g^{\ell}_{f,i}(x)$, which applies here because $g^{\ell_1}_{f,i}$ and $g^{\ell_2}_{f,i}$ tend to produce scores of different scales. Coefficients $\beta_1$ and $\beta_2$ are used to balance the weights from the two detectors. As shown in Figure~\ref{fig:coefficients} which is the result of a preliminary experiment, setting $\beta_1=\beta_2=0.5$ produces a close to optimal precision on all the given OOD, AD and NS data sets when CIFAR-10 is ID in a ResNet model. This figure also suggests that relying only on the MDL layer may provide acceptable results on detection of Out-of-Distribution anomalies (by treating TinyIm, LSUN, iSUN, SVHN as OOD) and noise detection (e.g., Gaussian noises), while relying only on the logit layer may provide acceptable results on detection of a few adversarial attacks. 

\paragraph{\textbf{Threshold.}} Similar to other methods~\cite{baseline,ODIN}, we need to have thresholds to distinguish normal inputs and anomaly inputs. Different from most other works, we define multiple thresholds based on classes of training samples. In the case of MNIST, there are $10$ classes of ID data, so we define a threshold for each class. When a sample $x$ is given to the DNN model which generates output class $i$, our approach collects data from the MDL layer and logit layer for the SVDDs to generate a score to compare with $\tau_i$. The threshold $\tau_{i}$ is computed in the way to ensure that $95\%$ of the testing samples from class $i$ of ID have scores above $\tau_{i}$. The threshold-based discriminator can be formally described as follows.

\begin{equation}\label{threshold}
    isAnomaly(x) = \left\{
    \begin{aligned}
     & True, \;\;\; if \;\; g_i(x) <  \tau_{i} \\
     & Fasle, \;\;\; if \;\; g_i(x) \geq \tau_{i}
    \end{aligned}
    \right.
\end{equation}

Note that SVDD performs poorly on high-dimensional data. Given the MDL layer is usually an early convolutional layer, the feature space of the MDL is often high-dimensional. In this case, we compute the mean of each channel to reduce the dimension of the extracted features from the MDL layer. More precisely, let $f^{\ell} \in \mathbb{R}^{d \times w \times h}$ be the feature maps of a convolutional layer, where $d$, $w$ and $h$ are depth, width and height, respectively. Then the feature size for the SVDD classifier is $d$, which is reduced from $d \times w \times h$ to $d$, 
with each dimension taking the average of all $w \times h$ values of the same depth.

\paragraph{\textbf{Nomalization.}} As we mentioned before the SVDD classifiers for the MDL layer and the logit layer tend to generate scores of different scales. If we simply combine the scores with incomparable scales, it is likely to weight one SVDD  classifier more than the other, leading to undesirable results. Therefore, a normalization process is essential. In our approach, we apply the min-max normalization procedure.
\begin{equation}\label{Eq:nomalization}
  score_{i}^{*} = \frac{score_{i} - score_{min}}{score_{max} - score_{min}}
\end{equation}
where the $score_{min}$ and $score_{max}$ are the minimum and maximum of the score vector, respectively.

In summary,
our SVDD detectors are trained from ID data only, i.e., $g^\ell_{f,i}$ only depends on the feature space at 
layer $\ell$
of training inputs if $\C(x)=i$. Most of the ID data are wrapped inside the hypersphere decision boundary in the feature space, as defined by Eq.~(\ref{eq:svdd}), and the hyperparameter $\nu$ controls the relative size $R$ and the percentage of training data to be outside of the boundary. 
Through some initial experiments we empirically choose an early layer, i.e., the MDL, and determine the coefficients $\beta_1$ and $\beta_2$, such that the detector $g_i(x) = \beta_1\cdot g^{\ell_1}_{f,i}(x)^* + \beta_2\cdot g^{\ell_2}_{f,i}(x)^*$ produces a score by combining information from the MDL (as $\ell_1$) and the logit layer (as $\ell_2$), if $x$ is likely to be from class $i$ (i.e., $\C(x)=i$). This score is then used to decide whether $x$ is anomaly.

\section{Experiments, evaluations and discussions}

We conduct experiment on three types of pre-trained DNN models, with three data sets chosen as ID, against various types of OOD, AD and NS data sets. Our testing code and data are publicly available at \url{https://github.com/fangzhenzhao/AnomalyDetection-keras}.

\begin{table*}[t]
    \caption{Comparison of our approach with the baseline, ODIN, Mahalanobis distance (MD) and ELO for OOD data.}\label{table:OOD}
    \addtolength{\tabcolsep}{-2pt}
    \begin{tabular}{cccc}
        \hline
        \multirow{2}{*}{Model} & \multirow{2}{*}{OOD} & TNR at 95\% TPR $\uparrow$  & AUROC $\uparrow$ \\ \cline{3-4}
        &  & \multicolumn{2}{c}{baseline / ODIN / MD / ELO / ours } \\
        \hline
         \multirow{2}{*}{\rotatebox{90}{\tabincell{c}{LeNet\\ (MNIST)} } }    & F-MNIST  &97.90 / 99.59 / 95.60 / 99.15 /  \textbf{99.72}\hspace{15pt}  & 99.26 / 99.81 / 98.81 / 99.77 / \textbf{99.83}  \\
                        & Omniglot   & 97.03 / 99.86 / 93.62 / \textbf{100.0} /  \textbf{100.0}   & 98.82 / 99.81 / 98.48 / \textbf{100.0} / 99.99 \\
                        \noalign{\vspace{4pt}}
        \hline

       \multirow{4}{*}{\rotatebox{90}{\tabincell{c}{VGG-16\\ (CIFAR10)} } }   &TinyIm    & 36.14 / 55.76 / 46.61 / 84.73 / \textbf{85.61}    & 89.79 / 92.79 / 90.68 / 96.45 / \textbf{97.54} \\
                        & LSUN    & 38.82 / 63.41 / 56.92 / 91.97 / \textbf{92.81}   & 90.90 / 94.58 / 93.05 / 98.13 / \textbf{98.98}  \\
                        & iSUN    & 39.06 / 63.65 / 54.90 / 89.22 / \textbf{92.64}   & 90.89 / 94.53 / 92.61 / 97.77 / \textbf{98.70}  \\
                        & SVHN    & 27.89 / 43.58 / 23.28 / \textbf{87.44} / 68.07    & 89.01 / 91.92 / 85.56  / \textbf{97.15} / 95.66  \\
        \hline
       \multirow{4}{*}{\rotatebox{90}{\tabincell{c}{ResNet\\ (CIFAR10)} } }  &TinyIm    & 30.32 / 48.41 / 15.52 / \textbf{95.08} / 94.24  &  87.13 / 90.58 / 79.46 / \textbf{98.91} / 98.61  \\
                        & LSUN    & 35.47 / 67.04 / 22.84 / 96.21 / \textbf{97.34}      & 89.66 / 94.72 / 85.22 / 99.07 / \textbf{99.25}   \\
                        & iSUN   & 34.25 / 63.47 / 22.45 / 93.61 / \textbf{95.34}      & 89.22 / 94.10 / 84.41 / 98.64 / \textbf{99.01}   \\
                        & SVHN    & 38.71 / 67.27 / 24.77 / 85.29 / \textbf{86.29}    & 89.74 / 93.80 / 82.19 / 96.74 / \textbf{97.00}     \\
       \hline

    \multirow{4}{*}{\rotatebox{90}{\tabincell{c}{VGG-16\\ (SVHN)} } }    & TinyIm    &  78.69 / 88.79 / 76.37 / 92.06 / \textbf{93.97}   & 96.92 / 97.99 / 96.56 / 98.21 /  \textbf{98.87}  \\
                        & LSUN    & 78.27 / 88.13 / 78.42 / 93.57 / \textbf{94.80}   & 96.80 / 97.83 / 96.71 / 98.39 / \textbf{99.05}   \\
                        & iSUN   & 81.48 / 90.83 / 79.82 / 93.55 / \textbf{95.38}   & 97.26 / 98.27 / 96.95 / 98.46 / \textbf{99.08}   \\
                        & CIFAR10  & 78.13 / 88.87 / 76.39 / 71.88 / \textbf{87.37}   & 97.67 / 97.91 / 85.69 / 94.62 / \textbf{97.83} \\
         \hline

  \multirow{5}{*}{\rotatebox{90}{\tabincell{c}{ResNet\\ (SVHN)} } }   & TinyIm      & 75.38 / 85.39 / 63.93 / 93.74 / \textbf{95.73}    & 96.21 / 97.17 / 94.37 / 98.62 / \textbf{99.01}  \\
                        & LSUN    & 72.67 / 83.02 / 58.96 / 96.27 / \textbf{96.58}   & 95.85 / 96.72 / 93.64 / 99.06 / \textbf{99.18} \\
                       & iSUN      &75.65 / 86.39 / 63.04 / 95.75 / \textbf{96.64}  & 96.28 / 97.30 / 94.20 / 98.99 / \textbf{99.21}  \\
                        & CIFAR10    & 74.12 / 85.12 / 63.91 / 80.14 / \textbf{87.60}   & 96.04 / 97.07 / 94.47 / 95.35 / \textbf{97.41}  \\
         \hline
    \end{tabular}
\end{table*}

\begin{table*}[h]
    \caption{Comparison of our approach with the KD+BU, LID, MD and ELO for adversarial data.}\label{table:AD}
    \addtolength{\tabcolsep}{-2pt}
    \begin{tabular}{cccc}
        \hline
        \multirow{2}{*}{Model} & \multirow{2}{*}{\tabincell{c}{AD}} & TNR at 95\% TPR $\uparrow$  & AUROC $\uparrow$ \\ \cline{3-4}
        &  & \multicolumn{2}{c}{KD+BU / LID / MD / ELO / ours } \\
        \hline
         \multirow{5}{*}{\rotatebox{90}{\tabincell{c}{LeNet\\ (MNIST)} } }   & FGSM    & 83.91 / 94.65 / 83.11 / \textbf{100.0} / \textbf{100.0} \hspace{15pt} & 86.86 / 89.03 / 96.66 / 99.97 / \textbf{99.98}   \\
                        & BIM-a   & \textbf{96.29} / 51.29 / 70.82 / 54.25 / 95.57   & 98.88 / 89.04 / 95.81 / 87.82 / \textbf{99.04}  \\
                        & BIM-b   & 28.51 / 34.70 / 23.63 / 99.28 / \textbf{99.32}   & 74.07 / 73.32 / 77.29 / 99.57 / \textbf{99.64}   \\
                        & JSMA    & \textbf{96.78} / 81.24 / 89.72 / 16.77 / 89.70       & \textbf{98.90} / 94.60 / 97.78 / 57.40 / 98.16  \\
                        & CW        & \textbf{97.23} / 60.30 / 66.33 / 6.77 / 79.73    & \textbf{99.07} / 90.60 / 95.12 / 50.81 / 97.12     \\
        \hline

       \multirow{5}{*}{\rotatebox{90}{\tabincell{c}{VGG-16\\ (CIFAR10)} } }   & FGSM  & 77.35 / 92.81 / 73.82 / 92.41 / \textbf{96.74}    & 87.66 / 98.11 / 95.75 / 98.59 / \textbf{99.22} \\
                        & BIM-a    & 40.58 / 7.76 / \textbf{85.93} / 1.47 / 63.57   & 76.36 / 64.40 / \textbf{97.27} / 51.08 / 94.10  \\
                        & BIM-b    & 11.76 / 3.44 / 2.34 / 1.21 / \textbf{15.80}     & 52.25 / 48.98 / 75.95 / 57.55 / \textbf{81.53}    \\
                        & JSMA   & 77.20 / 37.08 / \textbf{96.91} / 4.97 / 83.60    & 94.65 / 67.70 / \textbf{98.73} / 48.49 / 97.13    \\
                        & CW     & 49.92 / 6.93 / \textbf{86.22} / 7.39 / 81.18      & 80.04 / 66.11 / \textbf{96.96} / 65.27 / 96.53  \\
        \hline
       \multirow{5}{*}{\rotatebox{90}{\tabincell{c}{ResNet\\ (CIFAR10)} } }  & FGSM  & 38.93 / 99.95 / 9.70 / 99.98 /  \textbf{99.96}      & 75.70 / 99.98 / 82.14 / \textbf{100.0} / 99.81  \\
                        & BIM-a    & 17.48 / 11.74 / 6.63 / 14.56 / \textbf{35.65}       & 65.77 / 60.12 / 71.41 / 72.69 / \textbf{83.67}  \\
                        & BIM-b   & 99.89 / 80.43 / \textbf{99.99} / 95.17 / 99.98      & 99.93 / 90.74 / \textbf{99.99} / 98.66 / \textbf{99.99}   \\
                        & JSMA     & 60.0 / 71.80 / 35.00 / 83.76 / \textbf{91.62}     & 76.38 / 67.23 / 92.17 / 93.44 / \textbf{97.74}   \\
                        & CW       & 26.38 / 12.49 / 9.74 / 9.63 / \textbf{37.98}    & 75.75 / 60.91 / 82.99 / 63.21 / \textbf{86.84}    \\
       \hline

    \multirow{5}{*}{\rotatebox{90}{\tabincell{c}{VGG-16\\ (SVHN)} } }      & FGSM   & 86.91 / 99.90 / 83.50 / 99.91 / \textbf{97.69}         & 89.70 / 99.50 / 96.93 / 99.21 / \textbf{99.46}  \\
                        & BIM-a   & \textbf{57.05} / 15.64 / 53.43 / 4.21 / 40.81     & \textbf{91.19} / 74.40 / 89.81 / 55.35 / 81.36   \\
                        & BIM-b   & 1.35 / 76.36 / 22.69 / 20.09 / \textbf{98.84}     & 60.41 / 91.84 / 91.85 / 87.67 / \textbf{99.17}   \\
                        & JSMA    & \textbf{94.05} / 27.94 / 93.81 / 5.67 / 76.76       & 98.33 / 81.04 / \textbf{98.52} / 55.55 / 94.72   \\
                        & CW     & 75.20 / 11.28 / \textbf{100.0} / 4.43 / 66.88      & 97.03 / 71.18 / \textbf{98.48} / 49.38 / 94.54  \\
         \hline

  \multirow{5}{*}{\rotatebox{90}{\tabincell{c}{ResNet\\ (SVHN)} } }   & FGSM    & 73.89 / 99.56 / 72.93 / \textbf{100.0} / 99.76           & 83.01 / 99.55 / 95.30 / \textbf{99.85} / 99.80  \\
                        & BIM-a     & 50.60 / 16.81 / \textbf{54.5} / 3.44 / 24.53        & \textbf{89.66} / 74.60 / 89.40 / 43.52 / 77.33    \\
                        & BIM-b     & \textbf{100.0} / 99.12 /  \textbf{100.0} / 66.31 / \textbf{100.0}      &100.0 / 98.50 / \textbf{100.0} / 94.58 / 99.98  \\
                        & JSMA     & \textbf{84.33} / 34.73 / 84.05 / 8.28 / 51.04       &95.88 / 78.93 / \textbf{97.02} / 53.45 / 90.81  \\
                        & CW      & \textbf{83.17} / 9.26 / 78.72 / 4.48 / 54.53      & \textbf{97.26} / 63.81 / 96.68 / 48.40 / 92.36  \\
         \hline
    \end{tabular}

\end{table*}

\begin{table*}[thpb]
    \caption{Comparison of our approach with the baseline, ODIN and MD for noise data.}\label{table:NS}
    \centering
    \begin{tabular}{cccc}
        \hline
        \multirow{2}{*}{Model} & \multirow{2}{*}{NS} & TNR at 95\% TPR $\uparrow$  & AUROC $\uparrow$ \\ \cline{3-4}
        &  & \multicolumn{2}{c}{baseline / ODIN / MD / ours} \\
        \hline
         \multirow{3}{*}{\rotatebox{90}{\tabincell{c}{LeNet\\ (MNIST)} } }    & Gaussian  & 100.0 / 100.0 / 100.0 / \textbf{100.0}      & 99.58 / 100.0 / 99.77 / \textbf{100.0}  \\
                        & Uniform    & 99.27 / 100.0 / 99.25 / \textbf{100.0}     & 98.31 / 99.92 / 98.85 / \textbf{100.0} \\
                        & FoolIm    & 0.0 / 0.0 / 2.07 / \textbf{100.0}  & 75.56 / 86.73 / 68.61 / \textbf{99.74} \\
        \hline

       \multirow{3}{*}{\rotatebox{90}{\tabincell{c}{VGG-16\\ (CIFAR10)} } }   & Gaussian  & 7.75 / 58.42 / 97.43 / \textbf{100.0}       & 90.26 / 95.21 / 98.83 / \textbf{100.0}   \\
                        & Uniform     & 49.56 / 87.79 / 99.81 / \textbf{100.0}     & 94.72 / 96.78 / 99.28 / \textbf{100.0}   \\
                        & FoolIm   & 0.0 / 0.0 / 0.0 / \textbf{97.23}      & 71.60 / 79.53 / 71.82 / \textbf{98.14}   \\
        \hline
       \multirow{3}{*}{\rotatebox{90}{\tabincell{c}{ResNet\\ (CIFAR10)} } }  &Gaussian    & 30.22 / 58.51 / 33.67 / \textbf{100.0}       & 89.08 / 93.90 / 89.87 / \textbf{99.67} \\
                        & Uniform   & 24.56 / 53.40 / 23.48 / \textbf{100.0}       & 87.96 / 93.38 / 89.13 / \textbf{99.77}    \\
                        & FoolIm   & 0.0 / 0.0 / 16.83 / \textbf{100.0}      & 72.84 / 82.74 / 76.65 / \textbf{99.73}  \\
       \hline

    \multirow{3}{*}{\rotatebox{90}{\tabincell{c}{VGG-16\\ (SVHN)} } }    &  Gaussian   & 84.76 / 91.95 / 80.21 / \textbf{99.53}    & 97.67 / 98.49 / 97.00 / \textbf{99.70}  \\
                        & Uniform   & 90.49 / 96.35 / 83.88 / \textbf{98.99}     & 98.35 / 99.11 / 97.30 / \textbf{99.63}  \\
                        & FoolIm    & 0.0 / 0.0 / 0.0 / \textbf{100.0}   & 16.65 / 19.45 / 37.91 / \textbf{98.39}   \\
         \hline

  \multirow{3}{*}{\rotatebox{90}{\tabincell{c}{ResNet\\ (SVHN)} } }   & Gaussian  & 83.52 / 93.28 / 66.61 / \textbf{100.0}     & 97.28 / 98.32 / 95.31 / \textbf{99.92}  \\
                        & Uniform     &82.68 / 93.08 / 62.76 / \textbf{99.98}       & 42.09 / 46.74 / 94.89 / \textbf{99.86}  \\
                        & FoolIm    & 0.0 / 0.0 / 1.30 / \textbf{96.93}    & 42.09 / 46.74 / 53.18 / \textbf{98.63}   \\
         \hline
    \end{tabular}
\end{table*}

\subsection{Experiment settings}

We choose three popular DNN models used for image classification. All DNN models are pre-trained. The anomaly detection algorithm is run on a Windows $10$ desktop equipped with Intel I7-9700 $3.0$GHz processor, $16$G RAM and Nvidia GetForce GTX1660Ti.
\begin{enumerate}
\item A LeNet~\cite{mnist} model with two convolutional layers and three fully connected layers. The model is trained for the \textbf{MNIST} data set~\cite{mnist} and achieves $99.20\%$ accuracy on the testing set. \textbf{MNIST} consists of $60,000$ $28\times 28$ grayscale images of hand-written digits in the training sets and $10,000$ images in the testing set.
\item A ResNet~\cite{ResNet} model for the \textbf{CIFAR-10}~\cite{cifar10} data set and another ResNet model for the \textbf{SVHN}~\cite{svhn} data set, achieving accuracies of $91.65\%$ and $96.12\%$, respectively. \textbf{CIFAR-10} consists of $50,000$ and $10,000$ $32\times 32$ color images in its training set and testing set, respectively, with each image belonging to one of the ten classes. \textbf{SVHN} consists of $73,257$ and $26,032$ colored house numbers from Google Street View images in its training set and testing set, respectively.
\item A VGG~\cite{VGG} model for the \textbf{CIFAR-10} data set and another VGG model for the \textbf{SVHN} data set, achieving accuracies of $93.47\%$ and $95.56\%$, respectively.
\end{enumerate}

Outlier Exposure (OE) has been shown as an effective fine-tuning method for improving the performance of existing anomaly detectors~\cite{OutlierExposure,OECC,panda}. In this work, we also present the experimental results that combines OE and our approach, which are shown in the `ours + OE' column of Table~\ref{table:OOD_OE}, Table~\ref{table:AD_OE} and Table~\ref{table:NS_OE}. The authors of~\cite{OutlierExposure} demonstrate that only $50,000$ samples from auxiliary data set to be used to fine-tune the pre-trained model is enough to improve the performance of existing anomaly detectors. 
In this work, we use $50,000$ English letters from E-MNIST~\cite{EMNIST} as the auxiliary data set to fine-tune the LeNet model. For VGG and ResNet models, we use $50,000$ samples from the TinyImageNet data set~\cite{Imagenet} to perform the fine-tuning. Note that the training data from the auxiliary data set, the anomaly testing data and ID testing data are pairwise disjoint. 

\paragraph{\textbf{Evaluation Metrics.}}

Given a (binary) anomaly detector, we define true positive (TP) as the number of cases when an input from ID is correctly reported as $isAnomaly(x) = False$, and false negative (FN) as the number of cases when an input from ID is incorrectly reported as $True$, for anomaly. Similarly, true negative (TN) is the number of cases when an anomaly input is correctly reported as $True$, and false positive (FP) is the number of cases when an anomaly is incorrectly reported as $False$, for data from ID.
We adopt two commonly used metrics, TNR (True Negative Rate) at $95\%$ TPR (True Positive Rate) and Area Under the Receiver Operating Characteristic curve (AUROC), to evaluate the effectiveness of our method.

Since we have a detector $g_i$ for each class $i$, all counted values need to be taken weighted average for each class. For example, we have TPR = $\sum_{i=1,\dots d}\gamma_i\cdot \mbox{TPR}_i$, where TPR$_i$ is the true positive rate calculated for inputs that are classified as $i$ by the DNN and $\gamma_i$ is the percentage of sample cases being classified as $i$.

Existing works mostly focus on detecting either OOD only, or AD only. Therefore, we compare our results on OOD data sets and Noise data with models designed for OOD detection, and compare our results on AD data with models designed for adversarial attack detection, in separate.

\subsection{Experiment results}

\paragraph{\textbf{OOD Detection.}}


We consider several OOD data sets for evaluating the effectiveness of our methods. In particular, \textbf{Fashion-MNIST (F-MNIST)}~\cite{Fashion} and \textbf{Omniglot}~\cite{Omniglot} are used as OOD for the LeNet model trained with MNIST.  For the ResNet model trained with \textbf{CIFAR-10}, the OOD sets are \textbf{TinyImageNet} ~\cite{Imagenet}, \textbf{LSUN}~\cite{LSUN}, \textbf{iSUN} ~\cite{iSUN} and \textbf{SVHN} ~\cite{svhn}. For the ResNet model trained with \textbf{SVHN}, the OOD sets are \textbf{TinyImageNet}, \textbf{LSUN}, \textbf{iSUN} and \textbf{CIFAR-10}. The experiments for the two VGG models are treated in the same way as the ResNet models.
Note that we do not test MD with \emph{feature ensemble} which uses output from all layers but involves tuning with particular OOD sets. We only apply the version of MD which uses the logit layer of the DNN instead.


The results for OOD detection of these models are presented in Table~\ref{table:OOD}, where the data set enclosed by the brackets next to the model denotes the ID set, e.g., MNIST is the ID for the LeNet model. As shown in the results, our method has the best precision for detection of OOD anomalies in most cases. 
Even for the one case when ELO is better, the percentage difference is minor.


\begin{table*}[t]
    \caption{Comparison of results with OE about the baseline, ODIN, Mahalanobis distance (MD) and ELO for OOD data.}\label{table:OOD_OE}
    \addtolength{\tabcolsep}{-2pt}
    \setlength{\arraycolsep}{2pt}
    \begin{tabular}{cccc}

        \hline
        \multirow{2}{*}{Model} & \multirow{2}{*}{OOD} & TNR at 95\% TPR $\uparrow$  & AUROC $\uparrow$ \\ \cline{3-4}
        &  & \multicolumn{2}{c}{baseline+OE / ODIN+OE / MD+OE / ELO+OE / ours+OE } \\
        \hline
         \multirow{2}{*}{\rotatebox{90}{\tabincell{c}{LeNet\\ (MNIST)} } }    & F-MNIST  &98.98 / 99.49 / 94.85 / 99.13 / \textbf{99.73} \hspace{15pt} & 99.58 / 99.84 / 98.53 / 99.76 / \textbf{99.87}  \\
                        & Omniglot   & 99.91 / 99.97 / 98.82 / \textbf{100.0} / \textbf{100.0}    & 99.75 / 99.92 / 99.27 / \textbf{100.0} / \textbf{100.0}   \\
                        \noalign{\vspace{4pt}}
        \hline

       \multirow{4}{*}{\rotatebox{90}{\tabincell{c}{VGG-16\\ (CIFAR10)} } }   &TinyIm    & 55.74 / 66.46 / 65.85 / 81.24 / \textbf{81.93}    & 92.57 / 93.90 / 93.07 / 95.63 / \textbf{96.74}   \\
                        & LSUN    & 60.56 / 75.72 / 77.23 / 90.20 / \textbf{91.81}   & 94.02 / 95.98 / 96.10 / 97.68 / \textbf{98.66}  \\
                        & iSUN    & 60.07 / 74.49 / 75.26 / 87.91 / \textbf{90.20}    & 93.77 / 95.78 / 95.56 / 97.41 / \textbf{98.33}  \\
                        & SVHN    & 60.85 / 74.65 / 78.96 / 75.53 / \textbf{95.88}   & 94.80 / 95.92 / 94.58 / 95.94 / \textbf{98.22}  \\
        \hline
       \multirow{4}{*}{\rotatebox{90}{\tabincell{c}{ResNet\\ (CIFAR10)} } }  &TinyIm    & 46.50 / 65.42 / 13.90 / \textbf{94.43} / 93.55     & 90.78 / 93.60 / 80.95 / \textbf{98.76} / 98.47  \\
                        & LSUN    & 53.60 / 80.98 / 21.51 / 96.22 / \textbf{97.43}      & 93.02 / 96.74 / 86.31 / 99.06 / \textbf{99.27}   \\
                        & iSUN   & 53.23 / 80.94 / 21.69 / 93.74 / \textbf{95.55}      & 92.84 / 96.66 / 86.12 / 98.62 / \textbf{99.04}  \\
                        & SVHN    & 71.10 / 81.27 / 51.46 / 85.44 / \textbf{89.41}    & 95.61 / 97.35 / 91.22 / 96.78 / \textbf{97.66}     \\
       \hline

    \multirow{4}{*}{\rotatebox{90}{\tabincell{c}{VGG-16\\ (SVHN)} } }    & TinyIm    &  87.98 / 96.48 / 94.57 / 87.09 / \textbf{96.65}   & 97.78 / 99.17 / 98.54 / 97.36 / \textbf{99.30}  \\
                        & LSUN    & 85.50 / 95.48 / 93.33 / 84.92 / \textbf{95.75}   & 97.54 / 99.00 / 98.37 / 97.12 / \textbf{99.28}   \\
                        & iSUN   & 89.31 / 97.10 / 95.05 / 85.77 / \textbf{96.97}   & 97.96 / 99.26 / 98.70 / 97.29 / \textbf{99.35}    \\
                        & CIFAR10  & 84.54 / 92.22 / 93.56 / 59.61 / \textbf{92.30}   & 97.40 / 98.07 / 98.33 / 92.52 / \textbf{98.52}  \\
         \hline

  \multirow{5}{*}{\rotatebox{90}{\tabincell{c}{ResNet\\ (SVHN)} } }   & TinyIm      & 87.82 / 91.29 / 77.02 / 92.95 / \textbf{95.56}    & 97.52 / 98.10 / 95.75 / 98.45 / \textbf{98.91}  \\
                        & LSUN    & 86.20 / 90.38 / 74.55 / 95.79 / \textbf{96.68}   & 97.30 / 97.97 / 95.29 / 98.96 / \textbf{99.19}  \\
                       & iSUN      & 88.50 / 92.43 / 78.39 / 94.87 / \textbf{96.53}   & 97.66 / 98.33 / 95.82 / 98.84 / \textbf{99.08}  \\
                        & CIFAR10    & 88.41 / \textbf{91.95} / 78.26 / 78.98 / 87.94   & 97.59 / \textbf{98.27} / 95.89 / 95.10 / 97.35  \\
         \hline
    \end{tabular}
\end{table*}

\begin{table*}[h]
    \caption{Comparison of results with OE about the KD+BU, LID, MD and ELO for adversarial data.}\label{table:AD_OE}
    \addtolength{\tabcolsep}{-2pt}
    \begin{tabular}{cccc}
        \hline
        \multirow{2}{*}{Model} & \multirow{2}{*}{\tabincell{c}{AD}} & TNR at 95\% TPR $\uparrow$  & AUROC $\uparrow$ \\ \cline{3-4}
        &  & \multicolumn{2}{c}{KD+BU+OE / LID+OE / MD+OE / ELO+OE / ours+OE } \\
        \hline
         \multirow{5}{*}{\rotatebox{90}{\tabincell{c}{LeNet\\ (MNIST)} } }   & FGSM    & 78.21 / 95.85 / 86.18 / \textbf{99.99} / \textbf{99.99} \hspace{15pt}   & 93.91 / 95.93 / 97.23 / 99.93 / \textbf{99.98}   \\
                        & BIM-a   & 51.58 / 52.72 / 49.85 / 43.53 / \textbf{97.09}    & 88.31 / 87.49 / 91.59 / 84.01 / \textbf{99.29}  \\
                        & BIM-b   & 28.95 / 34.98 / 18.33 / \textbf{96.99} / 96.69    & 78.28 / 77.90 / 72.48 / 99.10 / \textbf{99.22}   \\
                        & JSMA    & 79.40 / 81.08 / 80.64 / 17.19 / \textbf{91.84}       & 95.12 / 92.89 / 96.69 / 58.80 / \textbf{98.42}  \\
                        & CW        & 46.74 / 37.65 / 37.62 / 6.21 / \textbf{92.46}    & 84.28 / 75.22 / 88.08 / 49.57 / \textbf{98.44}    \\
        \hline

       \multirow{5}{*}{\rotatebox{90}{\tabincell{c}{VGG-16\\ (CIFAR10)} } }   & FGSM  & 81.83 / 92.85 / 78.42 / 89.44 / \textbf{96.58}    & 66.79 / 98.73 / 96.07 / 97.99 / \textbf{99.14} \\
                        & BIM-a    & 73.95 / 12.55 / \textbf{76.07} / 1.34 / 67.55    & 78.86 / 67.37 / \textbf{96.17} / 49.81 / 95.24  \\
                        & BIM-b    & 5.77 / 8.47 / 7.03 / 1.05 / \textbf{10.63}     & 47.04 / 55.96 / 70.12 / 55.02 / \textbf{74.51}    \\
                        & JSMA   & 87.92 / 35.01 / \textbf{87.66} / 5.05 / 80.88    & 94.55 / 69.82 / \textbf{97.60} / 47.95 / 96.90    \\
                        & CW     & 75.53 / 10.33 / \textbf{77.29} / 4.79 / 74.75      & 80.02 / 67.87 / 96.27 / 63.06 / \textbf{96.36}  \\
        \hline
       \multirow{5}{*}{\rotatebox{90}{\tabincell{c}{ResNet\\ (CIFAR10)} } }  & FGSM  & 28.32 / 100.0 / 7.02 / \textbf{100.0} / \textbf{100.0}      & 75.40 / 99.99 / 80.22 / \textbf{100.0} / 99.79  \\
                        & BIM-a    & 12.14 / 11.12 / 5.78 / 18.05 / \textbf{38.84}       & 63.96 / 60.58 / 71.26 / 74.63 / \textbf{84.73}  \\
                        & BIM-b   & 99.84 / 66.52 / 99.96 / 95.93 / \textbf{99.98}     & 99.93 / 85.00 / \textbf{99.99} / 98.84 / \textbf{99.99}   \\
                        & JSMA     & 32.13 / 68.30 / 18.65 / 83.39 / \textbf{91.69}     & 73.45 / 67.78 / 89.57 / 93.36 / \textbf{97.83}   \\
                        & CW       & 13.11 / 11.23 / 7.19 / 13.98 / \textbf{48.12}    & 68.51 / 60.73 / 78.72 / 66.74 / \textbf{89.21}    \\
       \hline

    \multirow{5}{*}{\rotatebox{90}{\tabincell{c}{VGG-16\\ (SVHN)} } }      & FGSM   & 91.72 / \textbf{99.68} / 91.32 / 95.08 / 96.83         & 89.18 / \textbf{99.56} / 97.77 / 97.69 / 99.40  \\
                        & BIM-a   & 68.60 / 18.85 / 68.93 / 4.17 / \textbf{75.45}     & \textbf{94.03} / 75.77 / 93.83 / 53.17 / 93.10  \\
                        & BIM-b   & 15.26 / \textbf{57.61} / 21.04 / 8.63 / 52.43     & 69.16 / 88.40 / 85.54 / 81.62 / \textbf{93.30}   \\
                        & JSMA    & 80.22 / 28.44 / 78.11 / 4.76 / \textbf{91.64}       & 95.10 / 80.66 / 95.57 / 53.04 / \textbf{97.65}   \\
                        & CW     & 22.67 / 8.04 / 23.43 / 4.43 / \textbf{91.99}      & 78.16 / 57.97 / 76.28 / 49.18 / \textbf{97.97}  \\
         \hline

  \multirow{5}{*}{\rotatebox{90}{\tabincell{c}{ResNet\\ (SVHN)} } }   & FGSM    & 61.24 / 99.96 / 70.00 / \textbf{100.0} / 99.63          & 81.94 / \textbf{99.86} / 94.52 / 99.83 / 99.72  \\
                        & BIM-a     & 48.86 / 17.20 / \textbf{59.02} / 3.40 / 31.95        & 90.86 / 75.94 / \textbf{91.51} / 43.49 / 81.22    \\
                        & BIM-b     & 99.98 / 96.71 / 99.98 / 52.93 / \textbf{100.0}      & 99.99 / 97.54 / \textbf{100.0}/ 92.24 / 99.97  \\
                        & JSMA     & 68.19 / 31.36 / \textbf{73.28} / 8.00 / 58.82       & 94.08 / 79.76 / \textbf{95.59} / 52.52 / 92.05  \\
                        & CW      & 19.70 / 6.75 / 20.83 / 4.50 / \textbf{80.93}      & 80.88 / 58.54 / 81.35 / 48.71 / \textbf{96.53}  \\
         \hline
    \end{tabular}

\end{table*}

\begin{table*}[thpb]
    \caption{Comparison of results with OE about the baseline, ODIN and MD for noise data.}\label{table:NS_OE}
    \centering
    \begin{tabular}{cccc}
        \hline
        \multirow{2}{*}{Model} & \multirow{2}{*}{NS} & TNR at 95\% TPR $\uparrow$  & AUROC $\uparrow$ \\ \cline{3-4}
        &  & \multicolumn{2}{c}{baseline+OE / ODIN+OE / MD+OE / ours+OE } \\
        \hline
         \multirow{3}{*}{\rotatebox{90}{\tabincell{c}{LeNet\\ (MNIST)} } }    & Gaussian  & 100.0 / 100.0 / 100.0 / \textbf{100.0}      & 99.99 / 100.0 / 99.63 / \textbf{100.0}  \\
                        & Uniform    & 100.0 / 100.0 / 100.0 / \textbf{100.0}     & 99.95 / 99.99 / 99.59 / \textbf{100.0} \\
                        & FoolIm    & 30.57 / 51.29 / 30.44 / \textbf{100.0}  & 90.65 / 93.20 / 83.78 / \textbf{99.95} \\
        \hline

       \multirow{3}{*}{\rotatebox{90}{\tabincell{c}{VGG-16\\ (CIFAR10)} } }   & Gaussian  & 97.47 / 99.56 / 99.97 / \textbf{100.0}       & 98.38 / 98.79 / 99.42 / \textbf{100.0}   \\
                        & Uniform     & 97.82 / 99.60 / 99.96 / \textbf{100.0}     & 98.49 / 98.78 / 99.42 / \textbf{100.0}   \\
                        & FoolIm   & 2.92 / 5.96 / 11.14 / \textbf{98.37}      & 82.51 / 84.05 / 86.42 / \textbf{99.31}   \\
        \hline
       \multirow{3}{*}{\rotatebox{90}{\tabincell{c}{ResNet\\ (CIFAR10)} } }  &Gaussian    & 54.79 / 83.82 / 31.24 / \textbf{100.0}       & 93.58 / 97.28 / 91.33 / \textbf{99.76} \\
                        & Uniform   & 54.68 / 88.33 / 25.16 / \textbf{100.0}       & 93.77 / 97.62 / 90.77 / \textbf{99.73}    \\
                        & FoolIm   & 25.51 / 32.02 / 32.97 / \textbf{100.0}      & 84.59 / 86.04 / 85.34 / \textbf{99.86}  \\
       \hline

    \multirow{3}{*}{\rotatebox{90}{\tabincell{c}{VGG-16\\ (SVHN)} } }    &  Gaussian   & 99.19 / 99.91 / 99.42 / \textbf{99.98}    & 99.68 / 99.95 / 99.38 / \textbf{99.97}  \\
                        & Uniform   & 99.58 / 99.96 / 99.69 / \textbf{99.99}     & 99.83 / 99.98 / 99.54 / \textbf{99.97}  \\
                        & FoolIm    & 7.44 / 24.98 / 15.68 / \textbf{68.13}   & 63.92 / 69.08 / 62.91 / \textbf{96.25}   \\
         \hline

  \multirow{3}{*}{\rotatebox{90}{\tabincell{c}{ResNet\\ (SVHN)} } }   & Gaussian  & 94.31 / 96.49 / 83.93 / \textbf{100.0}     & 98.51 / 99.04 / 96.76 / \textbf{99.92}  \\
                        & Uniform     & 93.47 / 95.74 / 81.94 / \textbf{99.98}       & 98.43 / 98.89 / 96.57 / \textbf{99.89}  \\
                        & FoolIm    & 7.97 / 14.17 / 8.81 / \textbf{99.38}    & 75.51 / 78.40 / 74.01 / \textbf{99.11}   \\
         \hline
    \end{tabular}
\end{table*}

\paragraph{\textbf{AD Detection.}}

\begin{figure*}[th]
  \centering
  \includegraphics[scale=0.8]{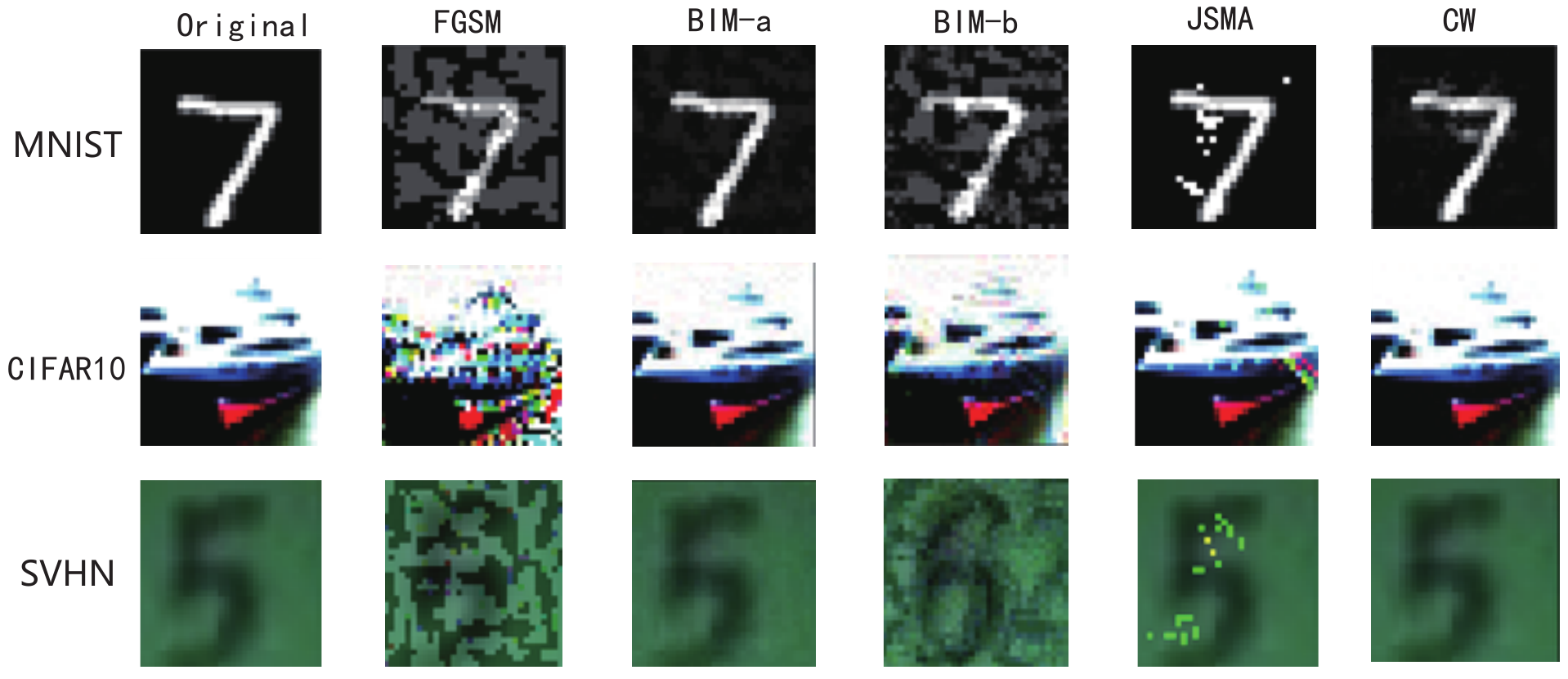}
  \caption{Some normal images and AD anomaly images for MNIST, CIFAR10 and SVHN.}\label{fig:adversarial samples}
\end{figure*}

We compare with the works that are designed for adversarial detection, including KD+BU~\cite{kd+bu}, LID~\cite{LID} and MD~\cite{Mahalanobis}.
We include the ELO method because this is the most related work to ours. The adversarial samples used in the experiment are generated by various well-known methods, including FGSM~\cite{FGSM}, BIM~\cite{KurakinGB16}, JSMA~\cite{PapernotMJFCS15} and CW~\cite{cw}.
For the BIM attack, we consider two scenarios: BIM-a, which stops iterating as soon as the attack is successful (‘at the decision boundary’), and BIM-b, which attacks for a fixed number of iterations that is well beyond the average misclassification point (‘beyond the decision boundary’). Some normal inputs and adversarial inputs are displayed in Figure ~\ref{fig:adversarial samples}.
Since both KD+BU and LID require training with adversarial inputs, in this experiment, both detectors are trained with FGSM.

The results in Table~\ref{table:AD} have shown that our method produces the best precision in about half of the cases regarding the AUROC values. For the rest cases where other methods have better precision, our scores are not much behind except for the two BIM-a cases for the SVHN data set. 


\begin{figure}[th]
  \centering
  \includegraphics[scale=0.9]{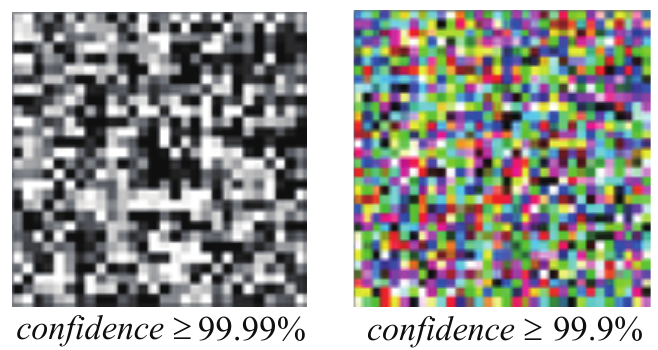}
  \caption{NS-\Romannum{2} anomaly inputs.}\label{fig:noise2}
\end{figure}

\paragraph{\textbf{NS Detection.}} In this experiment we have prepared three types of noise images.
\begin{enumerate}
\item The \textbf{Gaussian noise (NS-\Romannum{1})}  set consists of $10,000$ images of which every pixel is sampled from random Gaussian distribution with the mean $\mu = 0.5$  and the variance $\sigma = 1$, clipped to [0,1];
\item The \textbf{Uniform noise (NS-\Romannum{1})} set consists of $10,000$ images of which every pixel  is sampled from a random uniform distribution between $[0,1]$;
\item The \textbf{Fooling images (NS-\Romannum{2})} are generated by evolving meaningless images in order to mislead a DNN to output classes in ID with high confidence. We adopt the algorithm from Section~$3$ of~\cite{nguyen2015deep}. The Fooling Image sets feeding to the LeNet model consists of $10,000$ $28\times 28$ images (confidence$\geq 99.99\%$). The Fooling Image sets feeding to ResNet and VGG models both consist of $10,000$ $32 \times 32$ images (confidence$\geq 99.9\%$). Those images are totally unrecognizable to human eyes. Two examples from NS-\Romannum{2} are displayed in Figure ~\ref{fig:noise2}.

\end{enumerate}

The results shown in Table~\ref{table:NS} indicate that only our method achieves near $100\%$ precision regarding detection of the given Noise inputs. Note that some methods, including baseline~\cite{baseline} and ODIN~\cite{ODIN}, which take the classification probability of an input to discriminate whether this image is abnormal, 
fail to detect anomalies from NS-\Romannum{2} and in most cases classify them as ID with high confidence.
In particular, these methods mostly report $0$ in the column of TNR at 95\% TPR.
One possible explanation is that inputs from NS-\Romannum{2} are quite different from any known anomaly distributions that have been considered by these approaches, therefore, those images are relatively hard for them to~detect. 


%
%
%

\subsection{Comparison of results with Outlier Exposure}

Since OE is designed and proved to be effective for enhancing Out-of-Distribution (OOD) detection tasks~\cite{OutlierExposure}, we need to check if it is also useful in for our approach by using OE to fine-tune the models. However, in our experiment, it is inconclusive whether OE is an effective way to enhance precision when the target anomaly includes not only OOD, but also AD. First, out of the $18$ OOD detection benchmarks, OE has improved the results of our method for $11$ benchmarks regarding the AUROC metric. For adversarial (AD) detection tasks, our method combined with OE has produced equal or better performance in $15/25$ cases. Nevertheless, performance of our method has been enhanced in a few benchmarks, e.g., for the SVHN data set, the results of the two BIM-a are significantly improved ($11.74\%$ gain in VGG model and $3,89\%$ gain in ResNet model). For Noise (NS) detection, the performance of our method with OE is also improved in most cases (equal or better in $13/15$ cases), although the improvements are mostly minor.


Regarding the other methods, similar improvement in performance has been observed for basline, ODIN and MD regarding OOD detection tasks. For AD detection, the improvement of performance for KD+BU, LID, MD, ELO are in general insignificant (some cases even suffering performance degradation). The effect of OE for NS detection is in general positive for all the tested methods.

\subsection{Discussion}
As shown in Table~\ref{table:related-work}, both ODIN and MD apply \emph{input pre-processing} to improve their precision. To illustrate the performance of our work, we further compare our results with their best performance i.e., ODIN and MD with pre-processing, using VGG and ResNet models on the CIFAR-$10$ data set. The results (see Table~\ref{table:OOD2}) confirm that the performance of ODIN and MD with pre-processing is obviously better than those without pre-processing (Table~\ref{table:OOD}), and  demonstrate that our results are better than both ODIN and MD with pre-processing. Similarly, for AD detection, knowing the adversarial attack strategy, the performance of LID can be significantly improved and the performance of KD+BU can also be improved to some extent (excluding BIM-b for VGG model) as shown in Table~\ref{table:AD2}. More importantly, we show that our results outperform both KD+BU and LID with known adversarial samples in majority cases, when applied to VGG and ResNet models on the CIFAR-$10$ data set (see Table~\ref{table:AD2}).

\begin{table}[!th]
    \caption{The results of ODIN and MD for pre-processing.}\label{table:OOD2}
    \tiny
    \centering
    \begin{tabular}{cccc}
        \hline
         \multirow{2}{*}{Model} & \multirow{2}{*}{OOD}    & TNR at 95\% TPR $\uparrow$    & AUROC $\uparrow$ \\ \cline{3-4}
          &  & \multicolumn{2}{c}{ODIN / MD / ours  }  \\
        \hline

              \multirow{4}{*}{\rotatebox{90}{\tabincell{c}{VGG-16\\ (CIFAR10)} } }    &TinyIm    & 73.76 / 63.64 / \textbf{85.61}   & 95.66 / 92.14 / \textbf{97.54}   \\
         & LSUN       & 83.80 / 73.74 / \textbf{92.81}        & 97.36 / 94.87 / \textbf{98.98}     \\
        & iSUN         & 84.06 / 72.56 / \textbf{92.64}       & 97.27 / 94.51 / \textbf{98.70}   \\
               & SVHN         & \textbf{84.06} / 35.61 / 68.07  & \textbf{96.54} / 84.39 / 95.66   \\

     \hline
     \multirow{2}{*}{\rotatebox{90}{\tabincell{c}{ResNet\\ (CIFAR10)} } }  &TinyIm     & 65.35 / 41.32 / \textbf{94.24}    & 93.57 / 84.67 / \textbf{98.61}   \\
        & LSUN       & 85.40 / 58.12 / \textbf{97.34}         & 97.42 / 91.58 / \textbf{99.25}     \\
        & iSUN   & 82.33 / 56.00 / \textbf{95.34}         & 96.98 / 90.65 / \textbf{99.01}  \\
              &SVHN       & 67.27 / 60.26 / \textbf{86.29}           & 93.80 / 87.44 / \textbf{97.00}   \\
     \hline
    \end{tabular}

\end{table}

\begin{table}[!th]
    \caption{Comparison of our approach with the KD+BU, LID for the known adversarial data.}\label{table:AD2}
    \tiny
    \centering
    \begin{tabular}{cccc}
        \hline
        \multirow{2}{*}{Model} & \multirow{2}{*}{\tabincell{c}{AD}} & TNR at 95\% TPR $\uparrow$  & AUROC $\uparrow$ \\ \cline{3-4}
        &  & \multicolumn{2}{c}{KD+BU / LID / ours  } \\
        \hline

       \multirow{5}{*}{\rotatebox{90}{\tabincell{c}{VGG-16\\ (CIFAR10)} } }   & FGSM  &  77.35 / 92.81  / \textbf{96.74}     & 87.66 / 98.11 / \textbf{99.22}   \\
                        & BIM-a    & 86.38 / \textbf{82.79} / 63.57     & 83.74 / 84.15 / \textbf{94.10}    \\
                        & BIM-b    & 0.05 / \textbf{45.40} / 15.80     & 49.11 / \textbf{86.86} / 81.53    \\
                        & JSMA   & \textbf{97.23} / 94.81 /  83.60    & \textbf{97.29} / 97.15 / 97.13     \\
                        & CW     & 87.01 / \textbf{95.94} / 81.18   &  85.80 / 88.75 /  \textbf{96.53}    \\
        \hline
       \multirow{5}{*}{\rotatebox{90}{\tabincell{c}{ResNet\\ (CIFAR10)} } }  & FGSM  & 38.93 / 99.95 / \textbf{99.96}      & 75.70 / 99.98 / \textbf{99.81}   \\
                        & BIM-a    & 17.48 / \textbf{65.25} / 35.65    &  65.77 / 83.52 / \textbf{83.67}   \\
                        & BIM-b   & 99.89 / 99.89  / \textbf{99.98}        & 99.93 / 99.93 / \textbf{99.99}  \\
                        & JSMA     & 60.0 / 81.13 / \textbf{91.62}       & 76.38 / 90.69 / \textbf{97.74}   \\
                        & CW       & 26.38 / \textbf{49.92} / 37.98   & 75.75 / 72.55 / \textbf{86.84}    \\
       \hline

    \end{tabular}

\end{table}


\section{Conclusion and future work}

To enhance the applicability of DNN input anomaly detection in real-world tasks, we have proposed a novel approach that is able to detect all three types of anomalies, namely Out-of-Distribution (OOD) data, Adversarial (AD) data and Noise (NS) data. By combining the early and late layers of pre-trained DNN models, and deepening to a fine-grained level of each sub-class, our approach generally outperforms 
the state-of-the-art approaches for detection of all aforementioned anomaly types, to the best of our knowledge, which has been evidenced by the experiments.

One limitation is that this work and other anomaly detection works focus only on image classification. As the application domains of DNN is expanding fast, it is interesting and necessary to explore whether the existing 
methodology can be adopted to other applications beyond image processing, such as speech recognition, natural language processing, and intrusion detection with network traffic monitoring.

\bibliographystyle{sn-basic}
\bibliography{cas-refs}

\end{document}